\documentclass[runningheads]{llncs}
\usepackage[T1]{fontenc}
\usepackage{graphicx}

\usepackage{amsmath, amssymb, bm}

\usepackage{caption}
\usepackage{subcaption}

\DeclareMathOperator*{\argmin}{arg\,min}

\newfont{\bbb}{msbm10 scaled 700}

\newfont{\bb}{msbm10 scaled 1100}

\newcommand{\cv}{{\bf c}}

\newcommand{\lv}{{\bf l}}

\newcommand{\xv}{{\bf x}}
\newcommand{\yv}{{\bf y}}
\newcommand{\zv}{{\bf z}}

\newcommand{\etav}{\hbox{\boldmath$\eta$}}

\newcommand{\thetav}{\hbox{\boldmath$\theta$}}

\newcommand{\Lambdam}{\hbox{\boldmath$\Lambda$}}

\newcommand{\Phim}{\hbox{\boldmath$\Phi$}}

\usepackage[autostyle]{csquotes}

\usepackage[colorlinks = true,
            linkcolor = blue,
            urlcolor  = blue,
            citecolor = blue,
            anchorcolor = blue]{hyperref}

\usepackage{multirow}
\usepackage{ctable} 

\begin{document}
\title{A Deep-Discrete Learning Framework for Spherical Surface Registration}
%
%
\author{Mohamed A. Suliman \and
Logan Z. J. Williams \and
Abdulah Fawaz \and
Emma C. Robinson
}


\authorrunning{M. A. Suliman et al.}

\institute{Department of Biomedical Engineering,
King’s College London.\\
\email{\{mohamed.suliman, logan.williams, abdulah.fawaz, emma.robinson\}@kcl.ac.uk}}


%
\maketitle              
\begin{abstract}
Cortical surface registration is a fundamental tool for neuroimaging analysis that has been shown to improve the alignment of functional regions relative to volumetric approaches. Classically, image registration is performed by optimizing a complex objective similarity function, leading to long run times. This contributes to a convention for aligning all data to a global average reference frame that poorly reflects the underlying cortical heterogeneity. In this paper, we propose a novel unsupervised learning-based framework that converts registration to a multi-label classification problem, where each point in a low-resolution control grid deforms to one of fixed, finite number of endpoints. This is learned using a spherical geometric deep learning architecture, in an end-to-end unsupervised way, with regularization imposed using a deep Conditional Random Field (CRF). Experiments show that our proposed framework performs competitively, in terms of similarity and areal distortion, relative to the most popular classical surface registration algorithms and generates smoother deformations than other learning-based surface registration methods, even in subjects with atypical cortical morphology.

\end{abstract}

\section{Introduction}
\label{sec: intro}

The human cerebral cortex is highly convoluted structure, with complex patterns of functional organisation that vary considerably across individuals \cite{amunts2000brodmann,glasser2016multi}. Image registration is an important tool that supports comparison of brain images, through mapping of all data to a global average space. However, the degree of variation of cortical morphology and topography across individuals generates considerable uncertainty with regards to the optimal mapping between brains. 

Recently, cortical surface registration algorithms \cite{fischl1999high,yeo2009spherical,robinson2014msm,robinson2018multimodal} have led to improvements in the precision with which it is possible to compare features on the cortical surface through learning mappings that regularize deformations with respect to displacements along with the cortical sheet. Increasingly, these methods support multimodal registration \cite{robinson2014msm,robinson2018multimodal,zhao2021s3reg}, allowing improved evaluation of cortical functional areas \cite{coalson2018impact,glasser2016multi}. However, increasing evidence suggests that there is a limit to these improvements since cortical topographies vary in ways that break the diffeomorphic assumptions of classical registration algorithms \cite{glasser2016multi}.

To this end, learning-based registration algorithms \cite{dalca2019learning,heinrich2020highly} present an attractive framework for exploring these problems, on the grounds that they train fast, unsupervised deformation frameworks that can learn to adapt to sub-populations in the data \cite{dalca2019learning}. Previous learning-based registration frameworks have  predominately been generated for 2D or 3D Euclidean domains such as brain volumes \cite{balakrishnan2019voxelmorph,dalca2019learning,de2019deep,fan2018adversarial}, lung CT \cite{heinrich2019closing,heinrich2020highly,fu2020lungregnet}, and histology \cite{borovec2020anhir,shao2021prosregnet,pielawski2020comir}. However, increasing efforts have been made to adapt convolutional networks to non-Euclidean domains \cite{monti2017geometric,qi2017pointnet,zhao2019spherical}, resulting in the development of tools for learning-based registration of surfaces and point clouds \cite{aoki2019pointnetlk,wang2019deep,zhao2019spherical}. Most notably, the recent S3Reg framework \cite{zhao2021s3reg} proposes the first learning-based registration framework for cortical surfaces, leveraging the Spherical U-Net algorithm \cite{zhao2019spherical} to learn multi-resolution hexagonal filters across regular subdivisions of an icosphere.

One limitation of the Spherical U-Net is that its hexagonal convolutions do not generate a rotationally equivariant solution since the lack of a global coordinate system on a sphere causes filter orientation to flip at the poles. When learning registrations for S3Reg, this generates distortions that must be corrected through averaging warps learned across multiple rotated orientations of the sphere. Recent work showed that contrary to Spherical U-Net, MoNet convolutions (learned from a mixture of Gaussian kernels) could indeed be trained to be rotationally equivariant \cite{fawaz2021benchmarking}.  Therefore, we develop a new framework for spherical cortical registration based on MoNet, which also takes inspiration from deep-discrete registration frameworks \cite{heinrich2019closing,heinrich2020highly} designed to learn larger deformations than deep regression frameworks. We hypothesize that this could improve the generalization of our framework to brains with atypical topographies.

\noindent \textbf{Contributions} In this paper, we propose the first deep-discrete framework for cortical surface registration. We validate on the alignment of cortical folds and compare the proposed network against state-of-the-art classical \cite{fischl1999high,yeo2009spherical,robinson2014msm,robinson2018multimodal} and learning-based \cite{zhao2021s3reg} methods. Specific focus is placed on evaluating the smoothness of the generated warps and the capacity of the network to generalize to atypical cortical morphologies.

\section{Method}
\label{sec: method}
The proposed method combines ideas from the discrete frameworks of \cite{heinrich2020highly,robinson2014msm,robinson2018multimodal} to propose a Deep-Discrete spherical Registration (DDR) network for alignment of cortical surface features. The full DDR architecture consists of three networks that compose global rigid rotations and non-linear warps (learned in a coarse-to-fine fashion). The objective is to learn a spatial transformation $\Phim: M \to F$, that aligns cortical features on a moving mesh ($M$) to those of a fixed mesh ($F$) by optimizing a dissimilarity metric $\mathcal{L} $ of the form:
\begin{equation}
\hat{\thetav} = \argmin_{\thetav} \mathcal{L} \left(\Phim_{\thetav};F,M\right) + \Sigma\left(\Phim_{\thetav}\right),
\end{equation} 
where $\thetav$ are transformation parameters that parametrize $\Phim$, while $\Sigma\left(\cdot\right)$ is a regularization function that imposes smoothness on $\Phim$. 
 
Let $f\left(\cdot\right)$ represents our deep learning network architecture, with $\etav$ being a set of learnable parameters; then, our deep learning image registration problem may be represented as
\begin{equation}
\label{equ: img reg loss fun}
\thetav = f_{\etav}\left(F,M\right).
\end{equation}
In all cases, data is presented to the network as concatenated cortical metric maps of $F$ and $M$,  defined on a sphere $S^{2}$ that is parametrized by a sixth-order icosphere (that has 40962 vertices). The base architecture of each network is a MoNet U-Net. 

\subsection{Rotation Architecture}
Registration is learned in 3 stages, with the first network learning global rotations. To obtain the rotation matrix between $M$ and $F$, we use a MoNet U-Net to estimate the 12 parameters of the rotation matrix. With conventional rotation estimation methods, such as quaternions and Euler angles, being shown to be discontinuous in the real Euclidean spaces of four or fewer dimensions \cite{zhou2019continuity}, we apply the continuous 6D representation formula of the 3D rotations estimation proposed in \cite{zhou2019continuity}. The network optimization is driven using an unsupervised mean-squared-error loss. The rotationally aligned subjects are then passed to the next registration stage.

\subsection{Deep-Discrete Networks} 
\label{subsec: dir}
\begin{figure}[ht!]
\centering
\includegraphics[width=5 in]{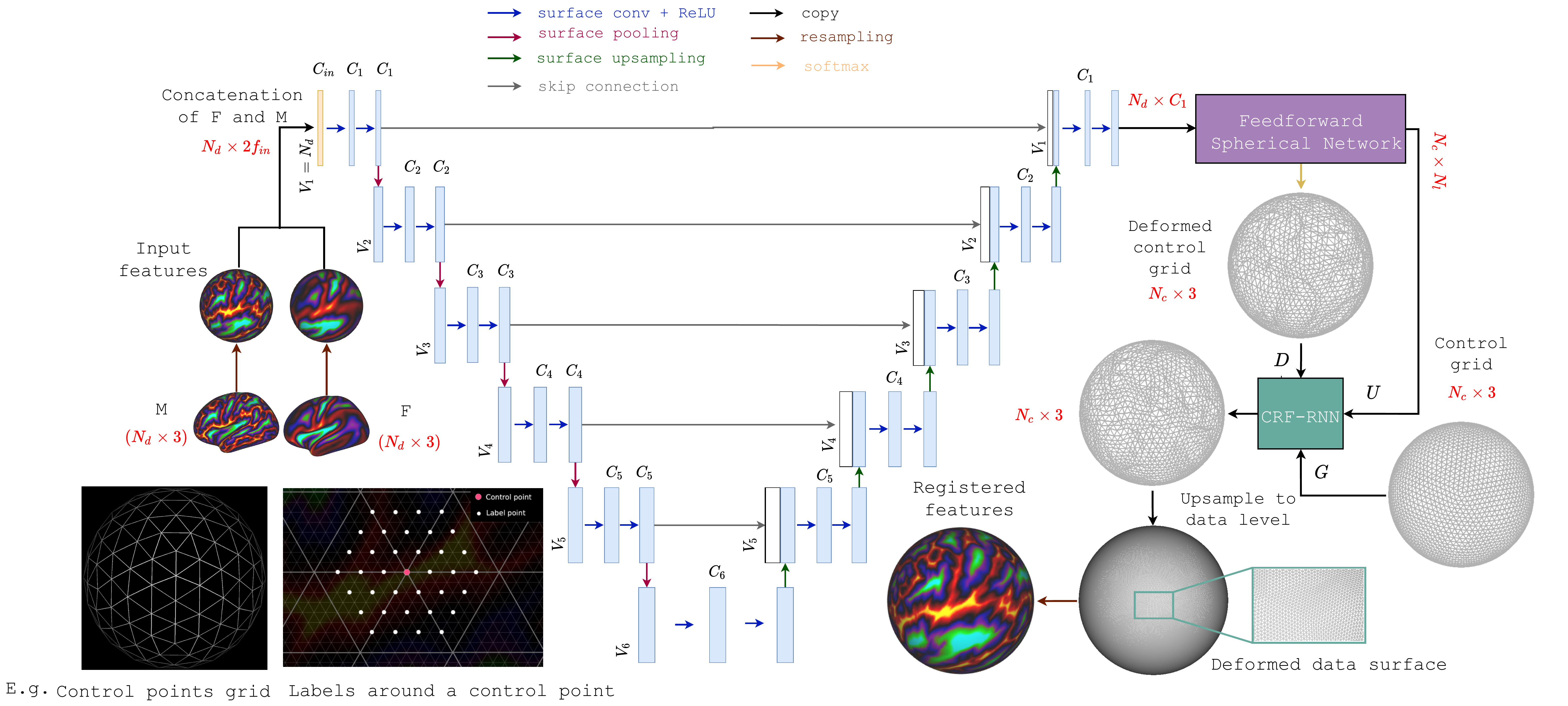}
\caption{DDR network architecture. The dimensions in red represent the input and the output dimensions at different levels on the network, while blue boxes represent features in the spherical space. $N_{d}$ is the number of data vertices, $N_{c}$ is the total number of control points,  $N_{l}$ is the number of labels around each control point, and $f_{in}$ is the number of features in each subject. } \label{fig: model}
\end{figure}

Deformable registration is subsequently learned using two deep-discrete networks, which learn optimal displacements as a classification problem, regularized by a deep CRF. Let $\{\cv_{i}\}_{i=1}^{N_{c}}  \in G \subset S^{2}$  be a set of $N_{c}$ control points, on the input image sphere, generated from a low-resolution icosphere (bottom left of Fig.~\ref{fig: model}), and let $\{\lv_{i}\}_{i=1}^{N_{l}} \in  S^{2}$ represent a set of label points, defined around each control point $\cv_{i}$, from a high-order icosphere (bottom left of Fig.~\ref{fig: model}). Then, the objective of DDR is to learn to predict the optimal label (displacement) for each control point to ensure features of the fixed and moving mesh optimally overlap. Importantly unlike classical discrete frameworks \cite{robinson2014msm,robinson2018multimodal}, for which run-time is linked to the label dimensionality, DDR is far less constrained by the extent of the label space.

Fig.~\ref{fig: model} shows an overview of our proposed approach. The first part of the network takes $M$ and $F$ concatenated to a single input, then passes this through a U-Net followed by a feedforward spherical network (FSN) to output $Q = \text{Softmax} \left(U\right) \in \mathbb{R}^{N_{c} \times N_{l}}$: softmax probabilities for each label. The second part of the network is a CFR-RNN network, which imposes smoothness by encouraging neighboring control points to take similar labels. 
\subsubsection{Classifier Architecture} The surface U-Net implements a symmetric encoder-decoder architecture with six resolution stages $i$, each defined at a different icosphere order. The number of icosphere vertices $V_{i}$ at each stage is related to the previous one through $V_{i+1} = \left(V_{i}+6\right)/4$. MoNet convolutional filters \cite{monti2017geometric} with a kernel size of 3, spherical polar pseudo-coordinates, and mean aggregation operators are applied to learn $C_{i}$ features at each level. In our network, we set $C_{i+1}=2C_{i}$ for all $i$. Each convolution is followed by a LeakyReLU activation with parameter 0.2. Downsampling and upsampling are performed using the hexagonal mean pooling and transpose convolution operations proposed by \cite{zhao2019spherical}.

The learned features at the final stage of the decoder, which is of dimensionality of $N_{d} \times C_{1}$, are then passed to a feedforward spherical network that decompresses channel dimensions over $r$ convolutional layers, each learning $\bar{C}_{i}$ features, to return an output of resolution $N_{d} \times N_{l}$ (a label prediction for each location in the input mesh); then regularized by downsampling label predictions $U$ to the desired control point resolution of $N_{c} \times N_{l}$. Once optimal labels are determined from the softmax operation $Q$, we deform our control grid accordingly using the spherical coordinates of the labels, i.e., $\lv_{i}$.

\subsubsection{CRF-RNN Network:} On its own, the classifier is of limited use since cortical registration is an ill-posed problem with many possible solutions. Therefore, smoothness is imposed through a CRF-RNN, adapted from \cite{zheng2015conditional}, that optimizes the following CRF energy cost function:
\begin{equation}
\label{equ: crf cost function}
E = \sum_{i} Q_{\left(\cv_{i},\lv_{i}\right)} +  \sum_{i \neq j}  \varphi \left(\lv_{\cv_{i}},\lv_{\cv_{j}}\right).
\end{equation}
Here, $Q_{\left(\cv_{i},\lv_{i}\right)}$ represents the cost (likelihood) of deforming $\cv_{i}$ to $\lv_{i}$ while 
\begin{equation}
\label{eq: energy modelling}
\varphi \left(\lv_{\cv_{i}},\lv_{\cv_{j}}\right)= \mu \left(\lv_{i},\lv_{j}\right) K_{G}\left(\lv_{\cv_{i}},\lv_{\cv_{j}}\right),
\end{equation}
measures the pairwise cost of deforming $\cv_{i}$ and $\cv_{j}$ to the label points $\lv_{i}$ and $\lv_{j}$, respectively. Moreover, $\mu$ is a learnable label compatibility function that captures correspondences between different pairs of label points (i.e.,  penalizing the assignment of different labels to different control points with similar properties), and $ K_{G}$ is a Gaussian kernel of the form \cite{krahenbuhl2011efficient,zheng2015conditional}
\begin{equation}
\label{eq: gaussion kernel}
K_{G}\left(\lv_{\cv_{i}},\lv_{\cv_{j}}\right) = \omega\left(\cv_{i},\cv_{j}\right) \exp\left(-\frac{1}{2\gamma^{2}}\left(  \left(\lv_{\cv_{i}}-\lv_{\cv_{j}}\right) \Lambdam  \left(\lv_{\cv_{i}}-\lv_{\cv_{j}}\right) \right)\right),
\end{equation}
with $\lv_{\cv_{i}}$ being the spatial location of the deformed $\cv_{i}$, $\omega$ being learnable filter weights, $\Lambdam$ being symmetric, positive-definite, kernel characterization matrix, and $\gamma$ being a kernel parameter. Energy (\ref{equ: crf cost function}) is optimized using the Recurrent Neural Network (RNN) implementation of \cite{zheng2015conditional}, which is based on learning multiple iterations of a mean-field CRF.

\subsubsection{Final Deformed Grid and the Loss Function:} The updated deformed control grid from the CRF-RNN network is upsampled to the input data icosphere order using barycentric interpolation. Then, the moving image features are resampled to the deformed data grid and compared to those of the fixed image. Optimization is driven using an unsupervised loss function (\ref{equ: img reg loss fun}) that is a sum of the mean-squared error and the cross-correlation.  To allow for more user control over the balance between accurate alignment and smooth deformation, we add a diffusion regularization penalty on the spatial gradients of the $\Phim$ in the form $\lambda \left( \left|\bigtriangledown \Phim_{\xv} \right| + \left| \bigtriangledown\Phim_{\yv} \right| + \left|\bigtriangledown \Phim_{\zv} \right| \right)$. The hexagonal filter is applied to compute $\bigtriangledown$.

\subsubsection{Coarse to Fine Registration:} 
As with surface registration methods, we perform multi-stage registration in the form of coarse-to-fine using two DDR networks. The first network is trained to align image features using a coarse grid of control points. The deformed grid is then upsampled to the higher level using the hexagonal upsampling method and passed to the second network that uses a higher resolution control grid.

\section{Experiments}
\label{sec: experiments}
To validate our proposed framework, we conduct a series of experiments that register individual cortical surfaces to an atlas. In each case, the proposed framework was compared against classical surface registration methods: FreeSurfer \cite{fischl1999high}, Spherical Demons \cite{yeo2009spherical}, and Multimodal Surface Matching (MSM) \cite{robinson2014msm,robinson2018multimodal}, as well as the learning-based method S3Reg \cite{zhao2021s3reg}.

\subsubsection{Datasets and Preprocessing:}
Experiments were performed using cortical surface data collected as a part of the adult Human Connectome Project (HCP) \cite{glasser2013minimal} resampled to a regular icosphere of order 6  using barycentric interpolation. A total number of 1110 brain MRI scans were used in the experiments with 888-111-111 split for train-validation-test and  batch sizes of 1 for all. For simplicity, registration was driven solely using sulcal depth as a feature, i.e., $C_{in}=2$; however, we point out that the network can be straightforwardly adapted to accept multi-channel features.

\subsubsection{Implementation and Training:}
MoNet filters are implemented using the PyTorch Geometric library \cite{Fey/Lenssen/2019}, while optimization is performed using ADAM  \cite{kingma2014adam} with a learning rate of $10^{-3}$. During the coarse registration stage, we use $N_{c} = 162$ vertices formed from an icosphere of order 2 with a 26.7 mm distance between adjacent points. $N_{l} = 600$ labels are generated from an icosphere of order 5 with 3.7 mm neighboring distances. Finally, we set $C_{1}= 32, \gamma = 0.7, \lambda=1.5, r=1,$ and $\bar{C}_{1}=600$.  For the fine registration, a total of $N_{c} = 2542$ points, formed using the vertices of an icosphere of order 4 with 6.9 mm neighboring distances, are applied. $N_{l} = 1000$ label vertices are generated from an icosphere of order 8 with 0.4 mm neighboring distances. Finally, we let $C_{1}= 2, \gamma = 0.2, \lambda=0.6, r=5$, and set $[\bar{C}_{i}]_{i=1}^{5} = [8,16,64,128,1000]$. All networks are trained on 100 epochs and we report the network performance with the best validation score.

\subsubsection{Baseline Registration Methods:}
We validate against the official implementations of Spherical Demons (SD)\footnote{\href{https://github.com/ThomasYeoLab/CBIG}{https://github.com/ThomasYeoLab/CBIG}}, MSM Pair \footnote{Available through FSLv6.0}, MSM Strain\footnote{\href{https://github.com/ecr05/MSM_HOCR}{https://github.com/ecr05/MSM$\_$HOCR}}, and the Spherical U-Net multi-stage registration algorithm (S3Reg)\footnote{\href{https://github.com/zhaofenqiang/SphericalUNetPackage}{https://github.com/zhaofenqiang/SphericalUNetPackage}}. Freesurfer is run as part of the HCP Pipeline, so we obtain its result from there. 

To achieve fair comparison across all methods, we ran parameter optimization and report performance across all runs. For SD, we run an additional 11 experiments by setting the number of smooth iterations to take $\left[1,5,10\right]$ and the smoothing variance $\sigma_{x} $ to be $\left[1,2,6,10\right]$. For MSM Pair, we run 22 experiments modifying the regularization penalty $\lambda \in \left[0.0001,  0.2\right]$, while for MSM Strain, we run 22 experiments with $\lambda \in \left[0.0001,  0.9\right]$. For S3Reg, we ran 7 more experiments by varying smoothness regularization at different registration levels to take the sets $[2,5,6,8]$, $[2,10,12,20]$, $[2,10,12,14]$, $[2,5, 12,16]$, $[2,10,6,8]$, $[2,10,12,8]$, and $[2,5,6,16]$; in each case training for 100 epochs and reporting the performance of the network with the best validation score. As FreeSurfer is not tunable by the user, we only report its results with default parameterization. Note that all these methods register two surfaces at 4 levels of icosphere subdivisions (coarse to fine), with S3Reg having an additional spherical transform network that ensures a diffeomorphic registration.

\subsubsection{Evaluation Metrics:}
All methods are compared in terms of cross-correlation (CC) similarity, areal strain ($J$), shape distortion $R$, and run time. Here $J=\lambda_1/\lambda_2$ and $R=\lambda_1/\lambda_2$, where $\lambda_1$ and $\lambda_2$ represent the eigenvalues of the local deformation gradient $F_{pqr}$ estimated from the deformation of each triangular face, defined by vertices: $p,q, r$. $\log_2J$ is equivalent to areal distortion.


\section{Results} 
\begin{figure}[b!]
\centering
\includegraphics[width=3 in]{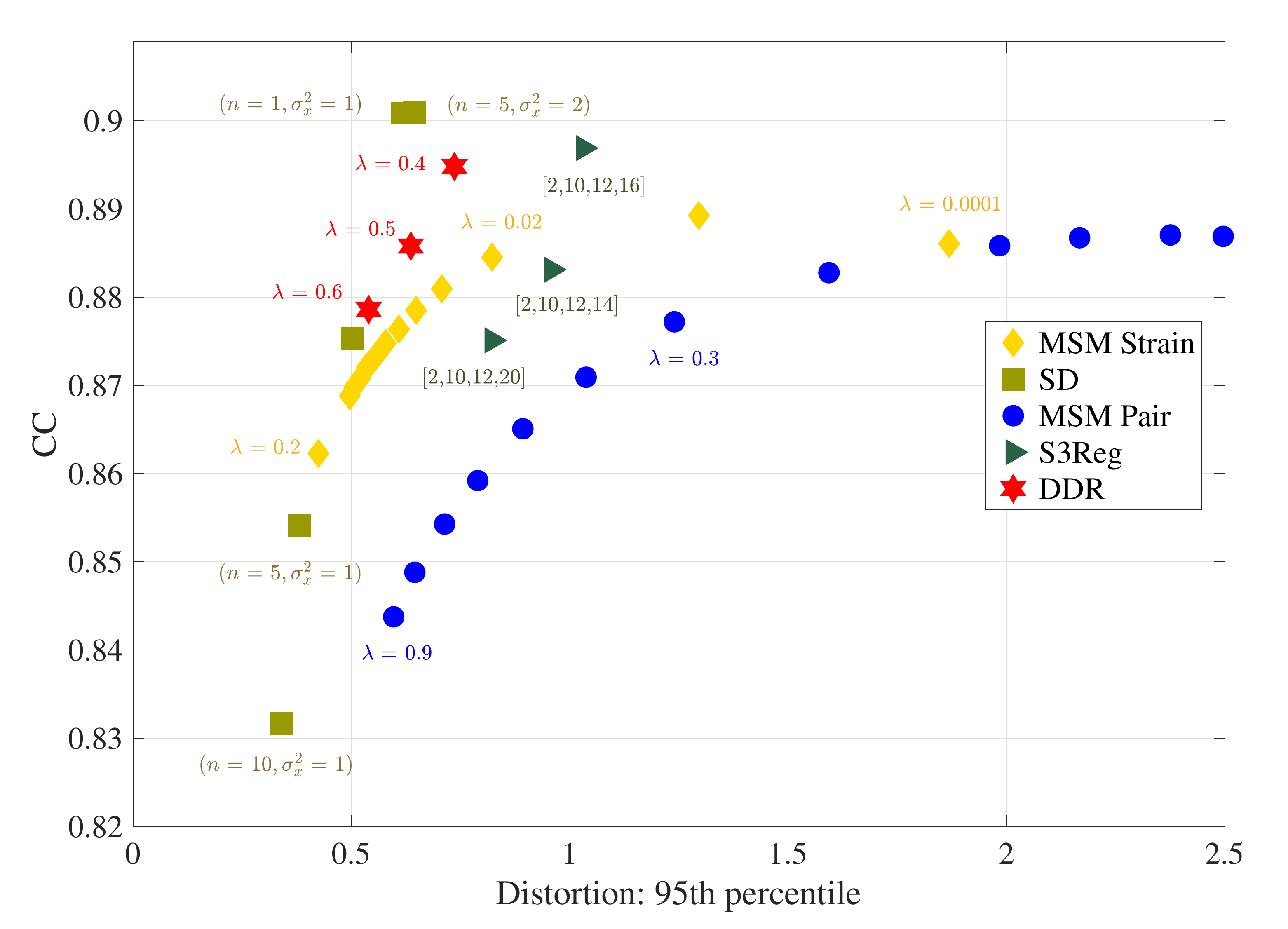}
\caption{Similarity performances of all methods vs. the 95th percentile of the areal distortion at multiple regularization levels across runs.}
\label{fig: strainj}
\end{figure}

Fig.~\ref{fig: strainj} plots the similarity performances of the different runs of all methods versus the 95th percentile of the absolute value of $J$. At each similarity level, DDR returns distortions within the range of the best classical methods (SD and MSM Strain) and reduced extremes of strain $J$ relative to that of S3Reg. 

\begin{figure}[h!]
     \centering
     \begin{subfigure}[b]{0.49\textwidth}
         \centering
         \includegraphics[width=\textwidth, height =1.72 in ]{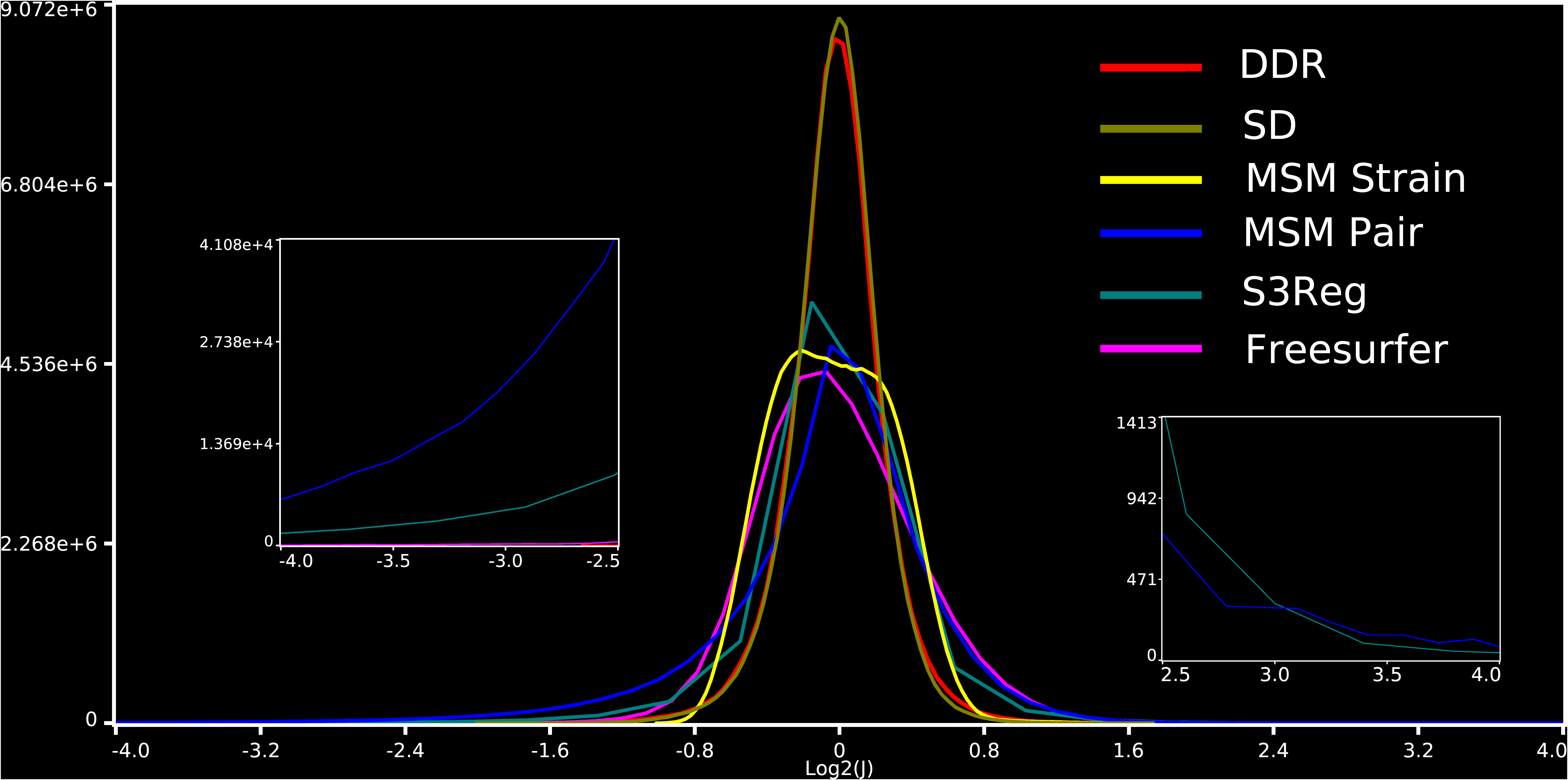}
         \caption{Strain $J$.}
         \label{fig}
     \end{subfigure}
     \begin{subfigure}[b]{0.49\textwidth}
         \centering
         \includegraphics[width=\textwidth, height =1.72 in]{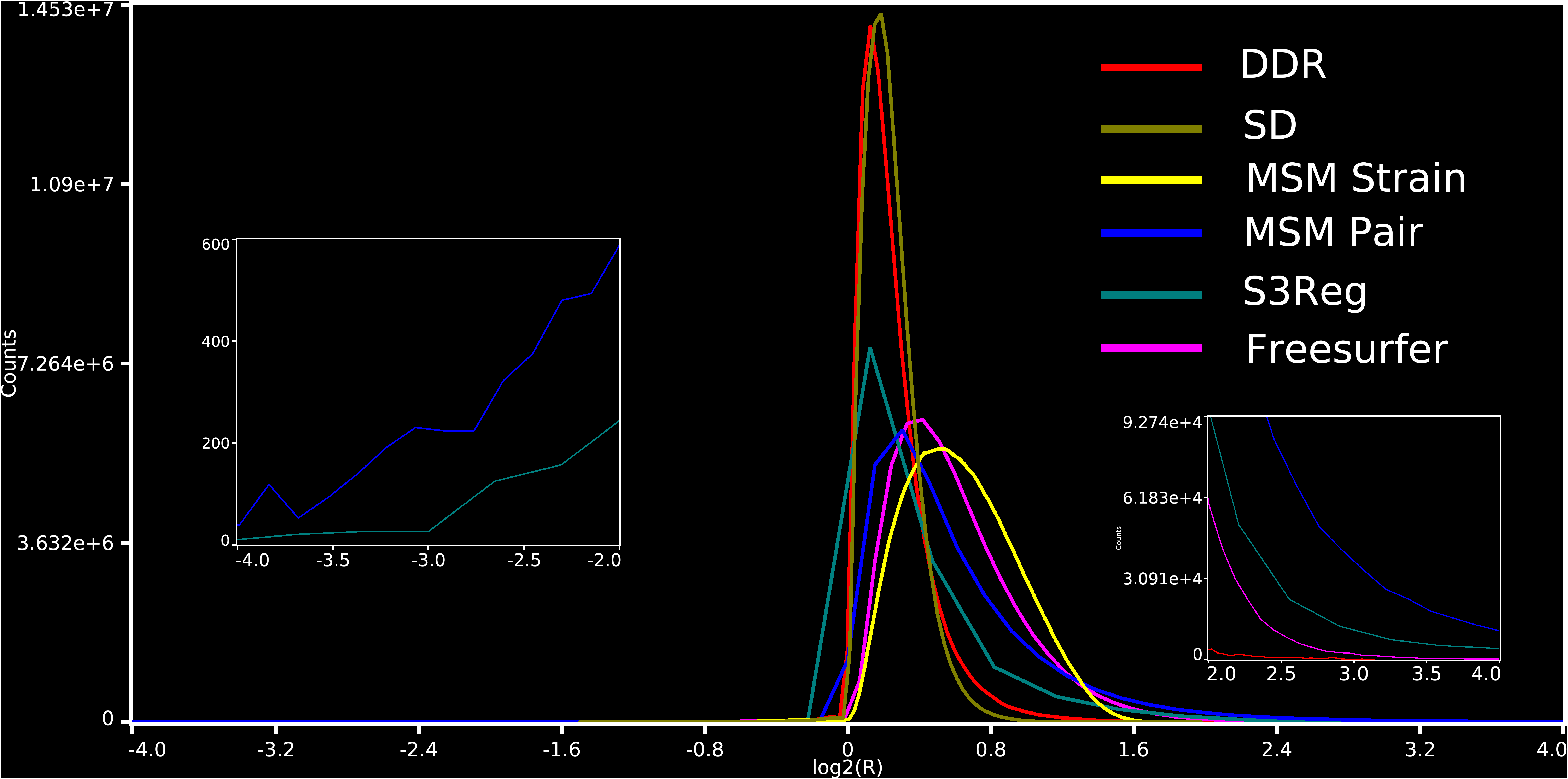}
         \caption{Strain $R$.}
         \label{fig}
     \end{subfigure}
     \caption{Histogram plots comparing methods distortions $J$ and $R$ across all test subjects for runs with CC $\sim 0.88$.}
     \label{fig: hist plot}
\end{figure}

This trend is repeated across the full distributions of $|J|$ and $|R|$ for each comparable run. Fig.~\ref{fig: hist plot} plots the full histogram distributions for $J$ and $R$ across all runs that return a CC value of approximately 0.88. Most of the areal distortions of SD and DDR at this CC level are around zero. However, S3Reg, MSM Pair, and Freesurfer all lead to extreme distortions across subjects, with an average of $150$ vertices at each subject returning $|J|$ above $2$ in S3Reg compared to $0.567$ for DDR.

Table.~\ref{table: dist} further reports summary statistics for these histogram distributions in terms of the mean, max 95th, and 98th percentiles of $|J|$ and the mean CC across all runs for these parameters. Again, DDR performs comparably to SD and MSM Strain, whereas S3Reg and MSM Pair provide the worst performance. These values reflect the best performance of S3Reg (in terms of distortion) across all runs. In terms of average run time, for a PC with NVIDIA Titan RTX 24GB GPU and Intel Core i9-9820X 3.30 GHz CPU, DDR has the least GPU and CPU times across all methods.

\begin{table}
\begin{center}
\hspace{-0.5cm}
\caption{Distortions measures and average runtime for different methods obtained for CC $\sim 0.88$; top: classical methods, bottom: learning-based methods.}\label{table: dist}\vspace{-7pt}
\centering
    \begin{tabular}{cc|c|c|c|c|c|c|c|c|}
        \hline
        \multicolumn{1}{|c|}{\multirow{2}{*}{\textbf{Methods}}}          & \multirow{2}{*}{\textbf{\begin{tabular}[c]{@{}c@{}}\textbf{CC}\\ \textbf{Similarity}\end{tabular}}} & \multicolumn{4}{|c|}{\textbf{Areal Distortion}}  & \multicolumn{2}{c|}{\textbf{Shape Distortion}} & \multicolumn{2}{c|}{\textbf{Avg. Time}}  \\ \cline{3-10}
        \multicolumn{1}{|c|}{}                 &           & \textbf{Mean}   & \textbf{Max}       & \textbf{95\%} & \textbf{98\%}                           &\textbf{Mean }      &\textbf{Max}       & \textbf{CPU}         &\textbf{GPU}  \\ \hline
        \hline
        \multicolumn{1}{|c|}{\begin{tabular}[c]{@{}c@{}}  Freesurfer  \end{tabular}}   
        
              &  0.75   & 0.34  & 11.73 & 0.82     &    1.00 & 0.63   &6.77  & 30 min   & -   \\

        \multicolumn{1}{|c|}{\begin{tabular}[c]{@{}c@{}} MSM Pair \end{tabular}}
        
         & 0.877        &0.41   & 9.17 & 1.24      &1.76  &0.62   & 9.05   &   13 min    & -     \\ 
     
     \multicolumn{1}{|c|}{\begin{tabular}[c]{@{}c@{}} MSM Strain \end{tabular}}
        
         & 0.880      & 0.27   &1.06  & 0.70       & 0.66    &0.64    &1.93  &  1 hour  & -        \\ 
         
         \multicolumn{1}{|c|}{\begin{tabular}[c]{@{}c@{}}  SD  \end{tabular}}   
        
        & 0.875 & 0.18 &  2.00  &0.50   &0.65 &0.24  &1.98 & 1 min  &  -    \\ \hline \hline

         \multicolumn{1}{|c|}{\begin{tabular}[c]{@{}c@{}}  S3Reg \end{tabular}}
        
         & 0.875      & 0.266   &22.22 & 0.82   & 1.16  &0.51    &  21.65  &      8.8 sec         &    8.0 sec    \\ 
                 
         \multicolumn{1}{|c|}{\begin{tabular}[c]{@{}c@{}} DDR \end{tabular}}
        
         & 0.878          & \textbf{0.19}  &\textbf{2.66}   &\textbf{0.53}  & \textbf{0.71} &\textbf{0.26 }  &\textbf{3.14}  & \textbf{ 7.7 sec }              &   \textbf{2.3 sec}           \\ \hline \hline
    \end{tabular} 
    \end{center}   
\end{table}

\begin{figure}[h!]
\begin{center}
\includegraphics[width=3.4 in]{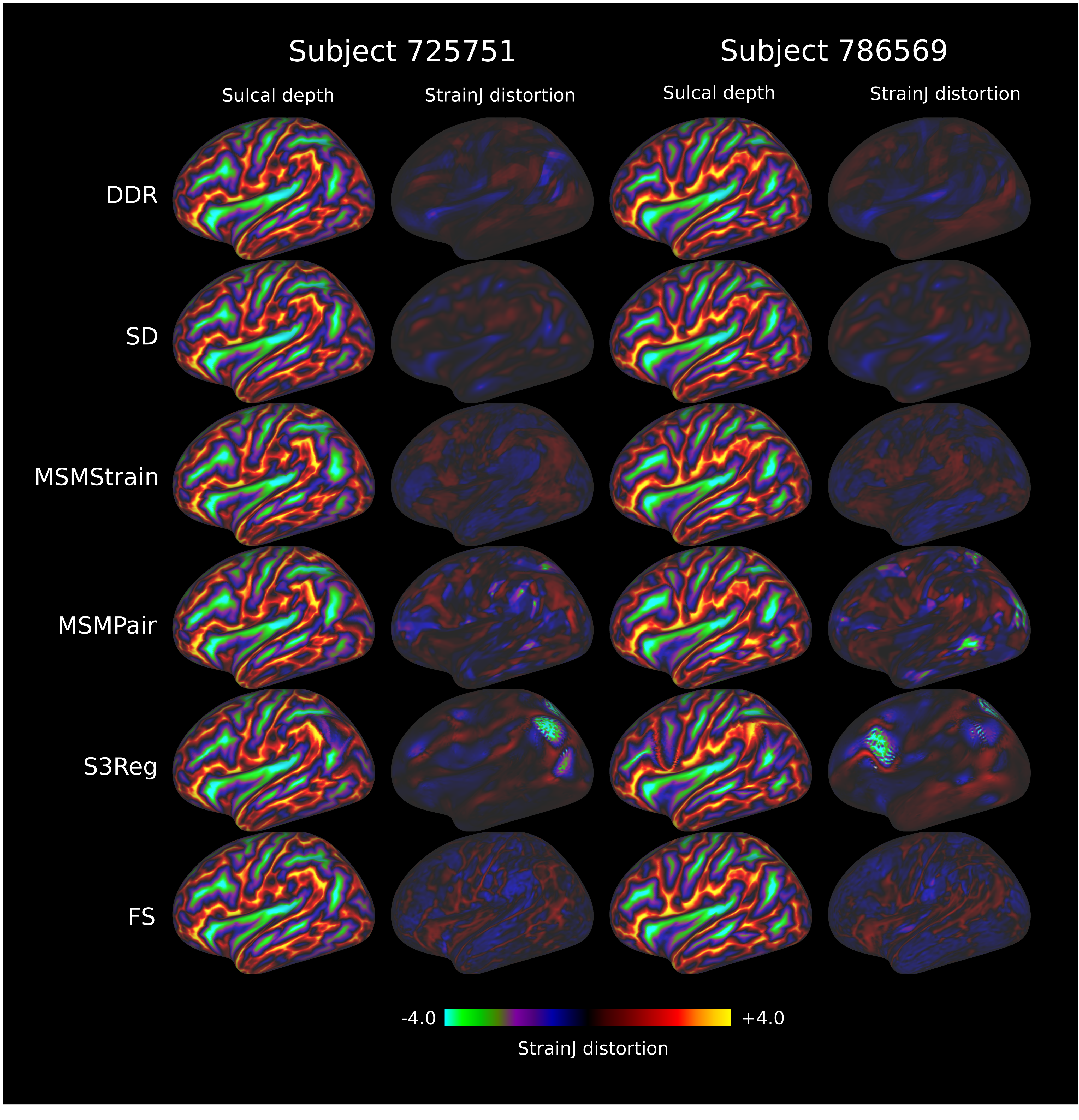}
\caption{Qualitative comparisons of sulcal depth alignment for 2 subjects. Results reflect distortion values presented in Fig.~\ref{fig: hist plot} and Table.~\ref{table: dist}. Columns 1 and 3 show registered sulcal depth maps; columns 2 and 4 reflect strain $J$ across the surface. Cyan areas of the distortion maps highlight significant distortions.} 
\label{fig: reg subjects}
\end{center}
\end{figure}

Fig.~\ref{fig: reg subjects} presents a qualitative comparison of the alignment quality for 2 subjects across all methods, together with the strain $J$ metrics across the surface. Results show that DDR, SD, and MSM Strain generate comparable alignment for reduced distortion relative to S3Reg and MSM Pair that produce alignments that result in peaks of high distortions.

\begin{figure}[h!]
     \centering
     \begin{subfigure}[b]{0.49\textwidth}
         \centering
         \includegraphics[width=\textwidth, height=1.8 in]{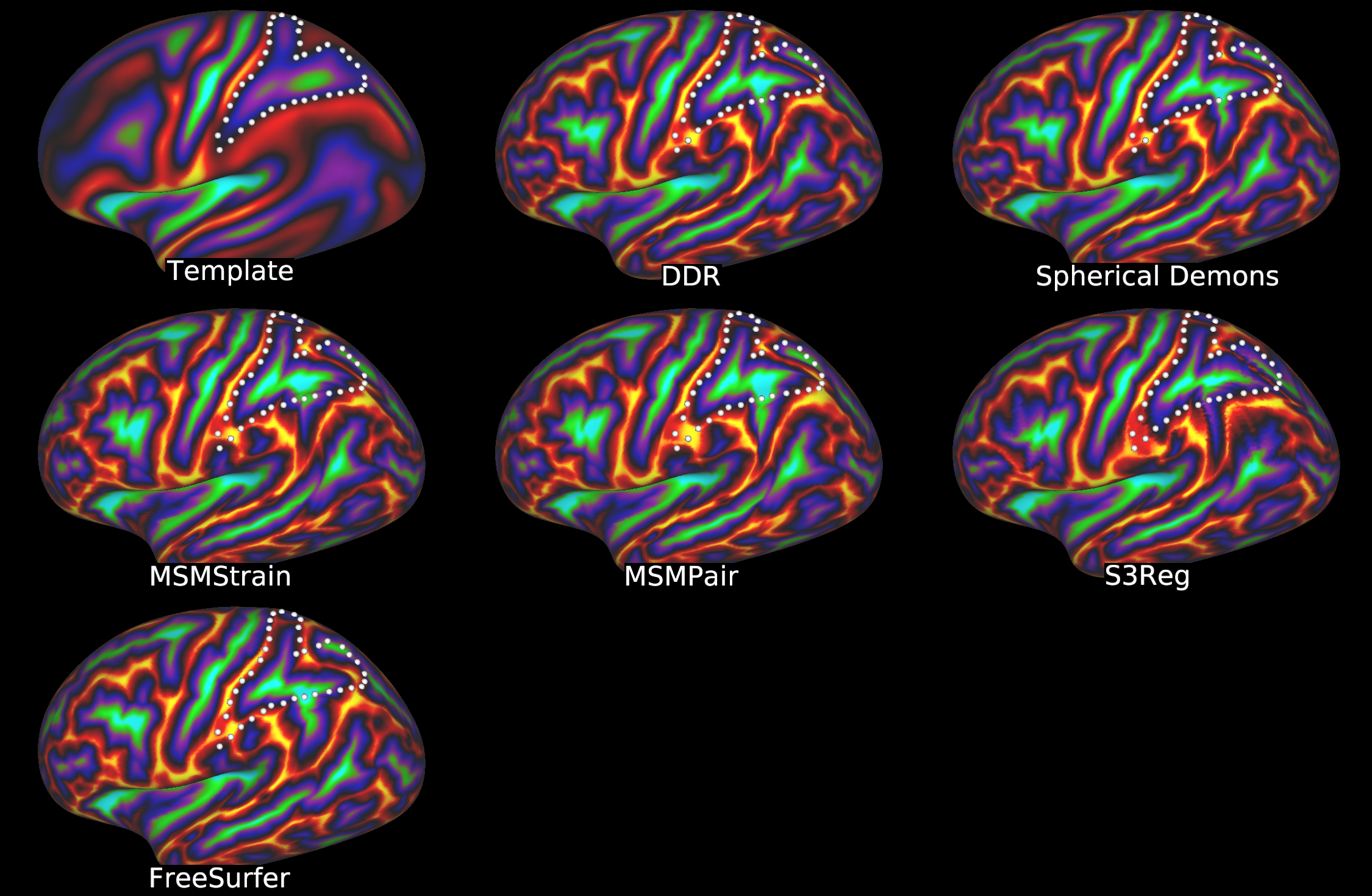}        
         \caption{}
         \label{fig: subject a}
     \end{subfigure}
     \begin{subfigure}[b]{0.49\textwidth}
         \centering
         \includegraphics[width=\textwidth]{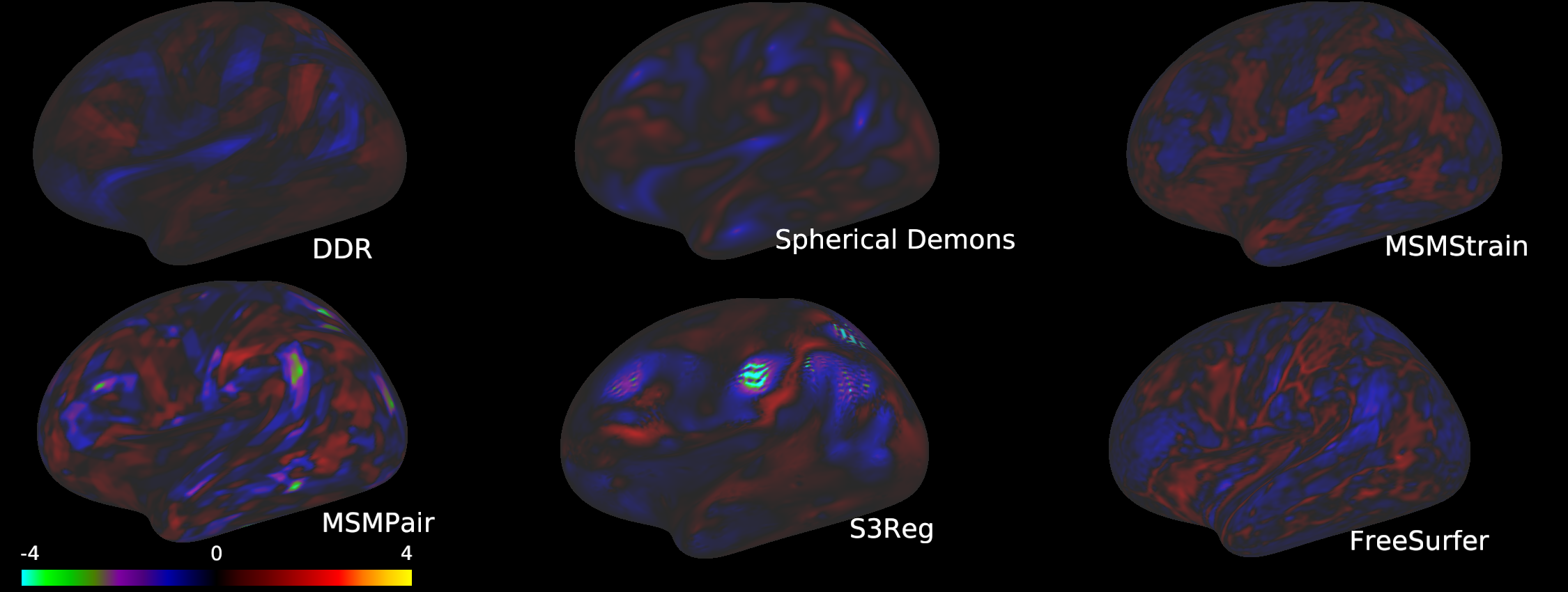}
         \includegraphics[width=\textwidth]{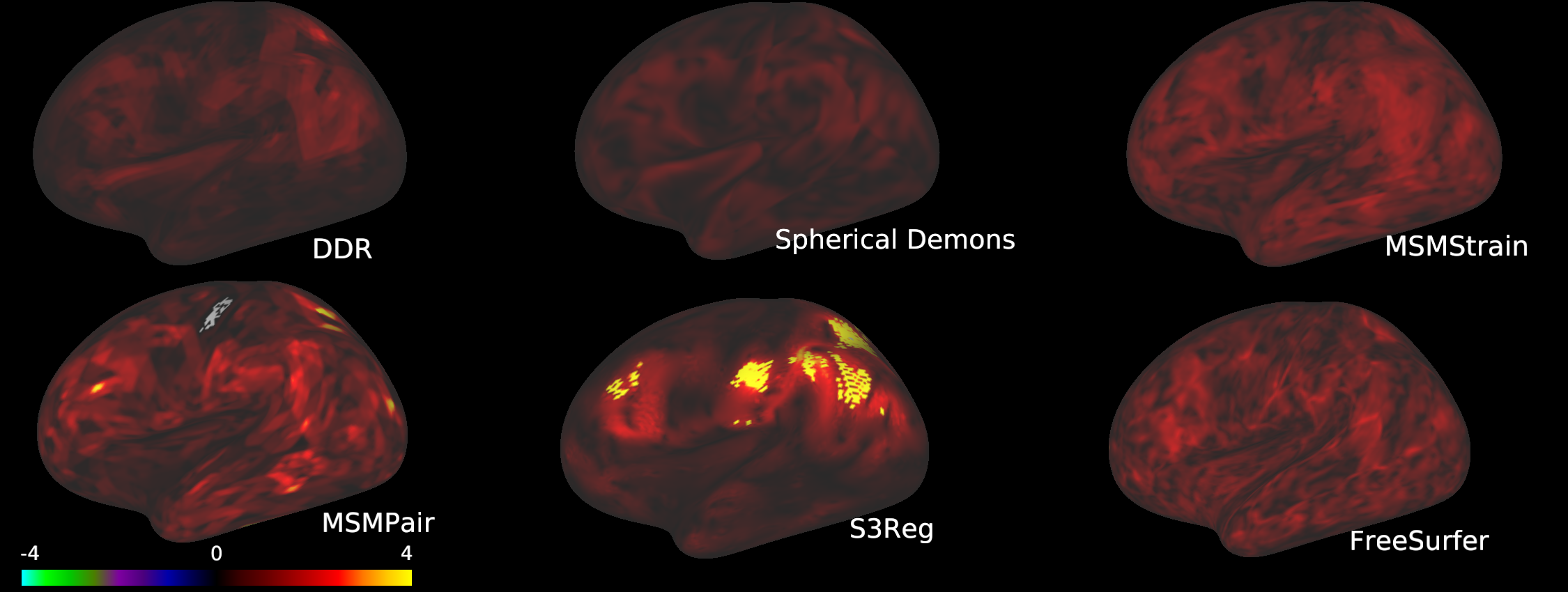}
         
         \caption{}
         \label{fig: subject b}
     \end{subfigure}
     
     \vspace{10pt}
     \begin{subfigure}[b]{0.49\textwidth}
         \centering
         \includegraphics[width=\textwidth, height=1.8 in]{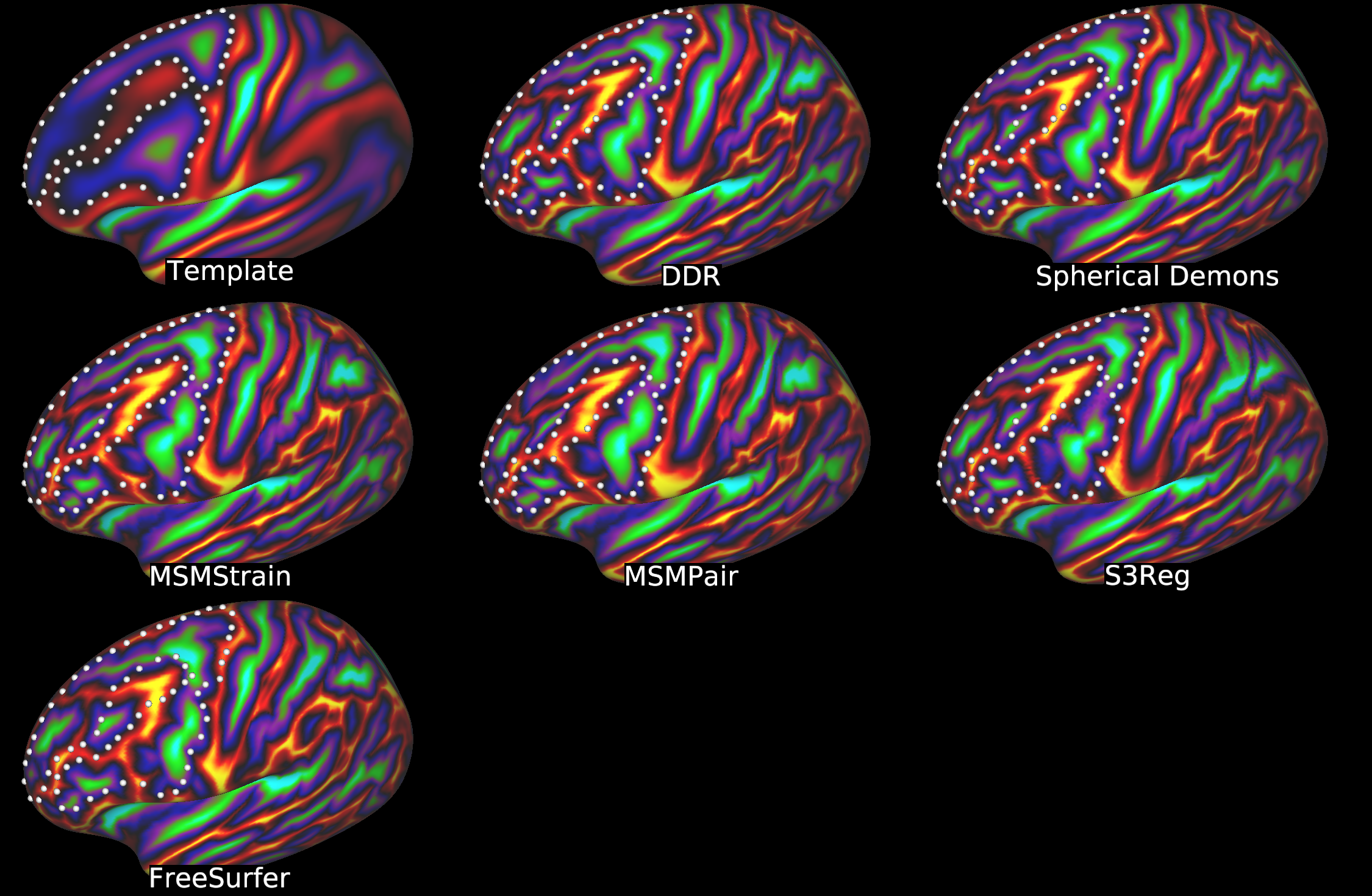}
         \caption{}
         \label{fig: subject a}
     \end{subfigure}
     \begin{subfigure}[b]{0.49\textwidth}
         \centering
         \includegraphics[width=\textwidth]{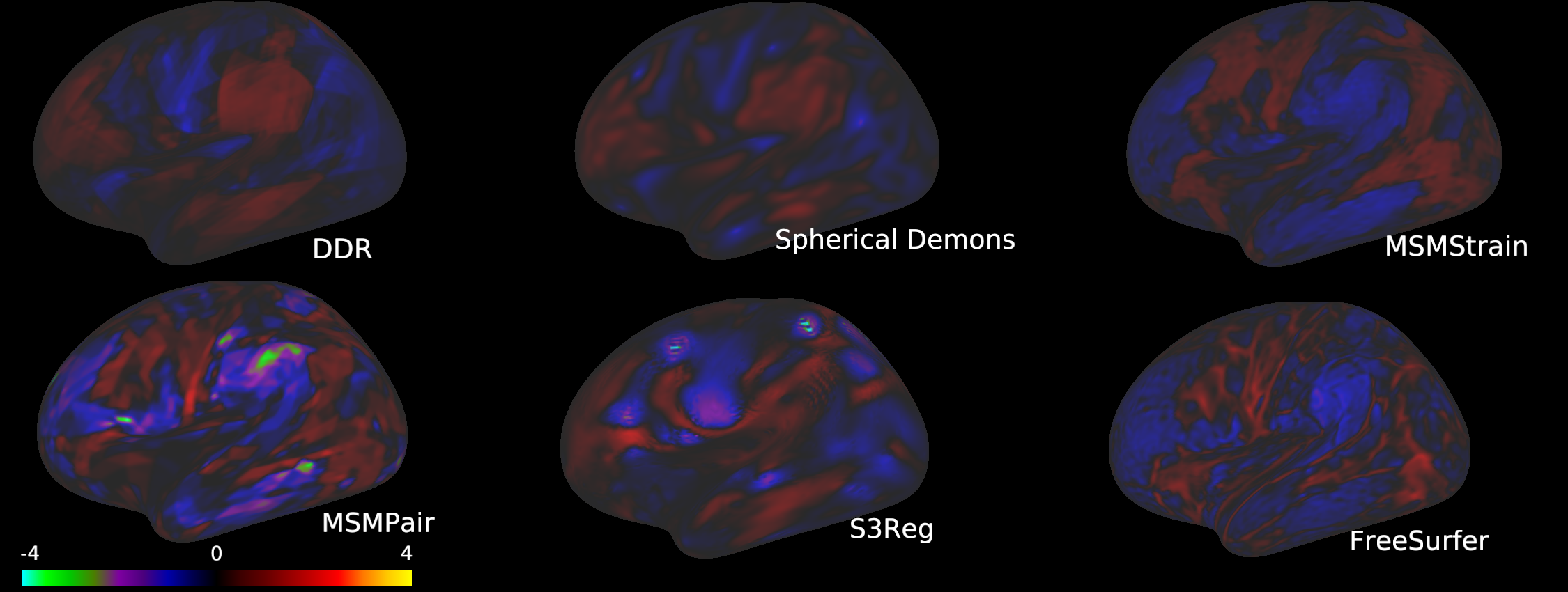}
         \includegraphics[width=\textwidth]{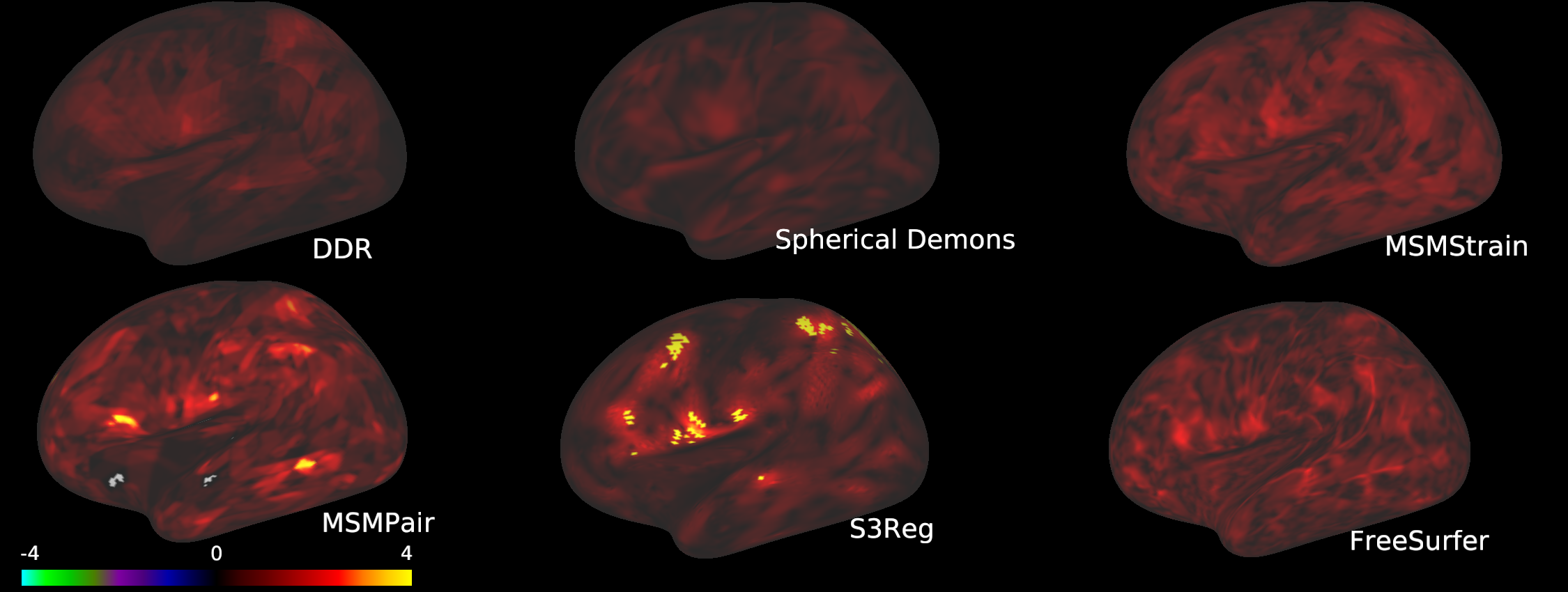}
         
         \caption{}
         \label{fig: subject b}
     \end{subfigure}
     
     \vspace{10pt}
          \begin{subfigure}[b]{0.49\textwidth}
         \centering
         \includegraphics[width=\textwidth, height=1.8 in]{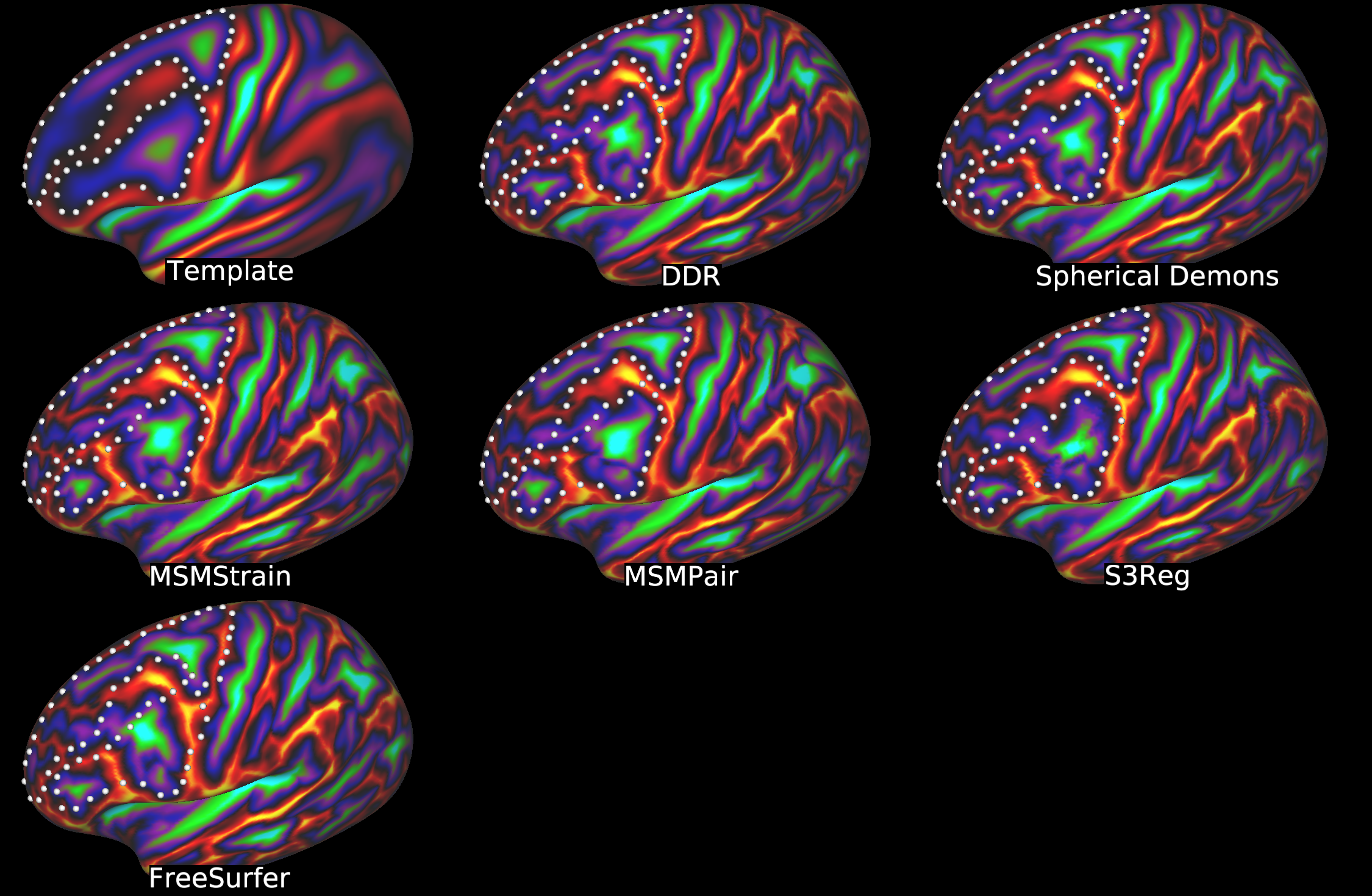}
         \caption{}
         \label{fig: subject a}
     \end{subfigure}
     \begin{subfigure}[b]{0.49\textwidth}
         \centering
         \includegraphics[width=\textwidth]{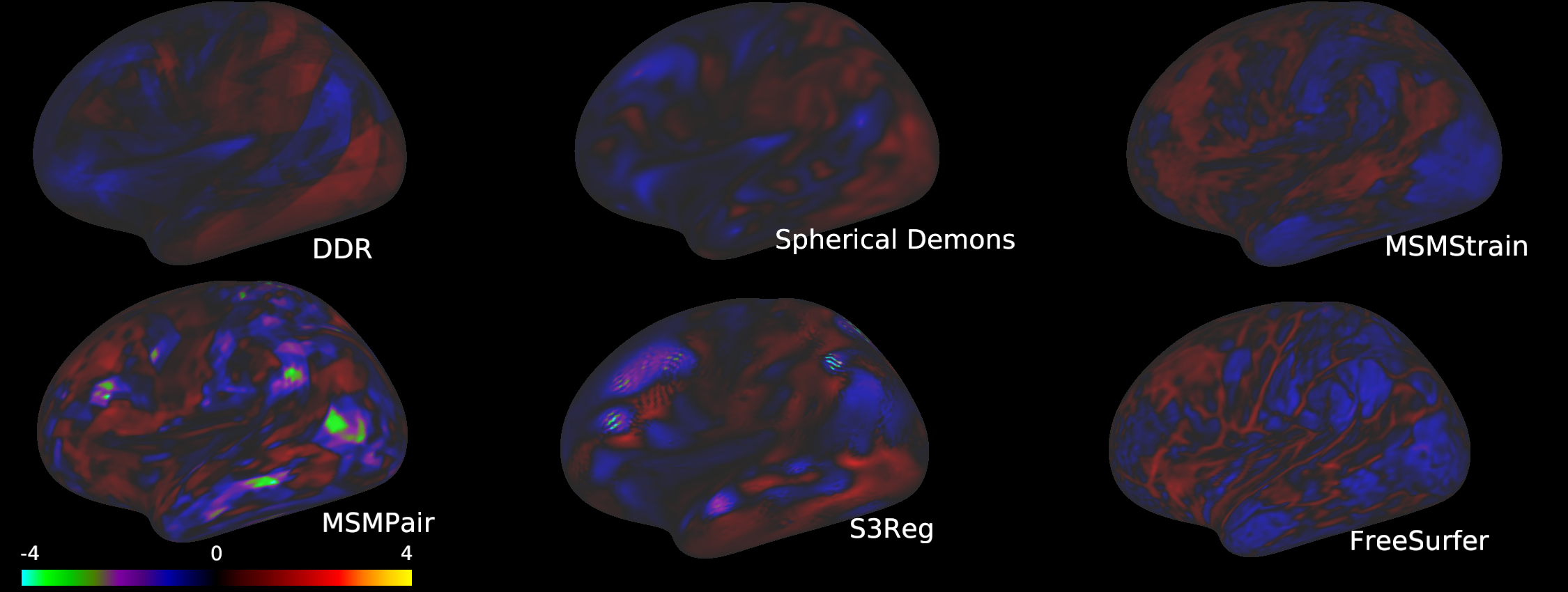}
         \includegraphics[width=\textwidth]{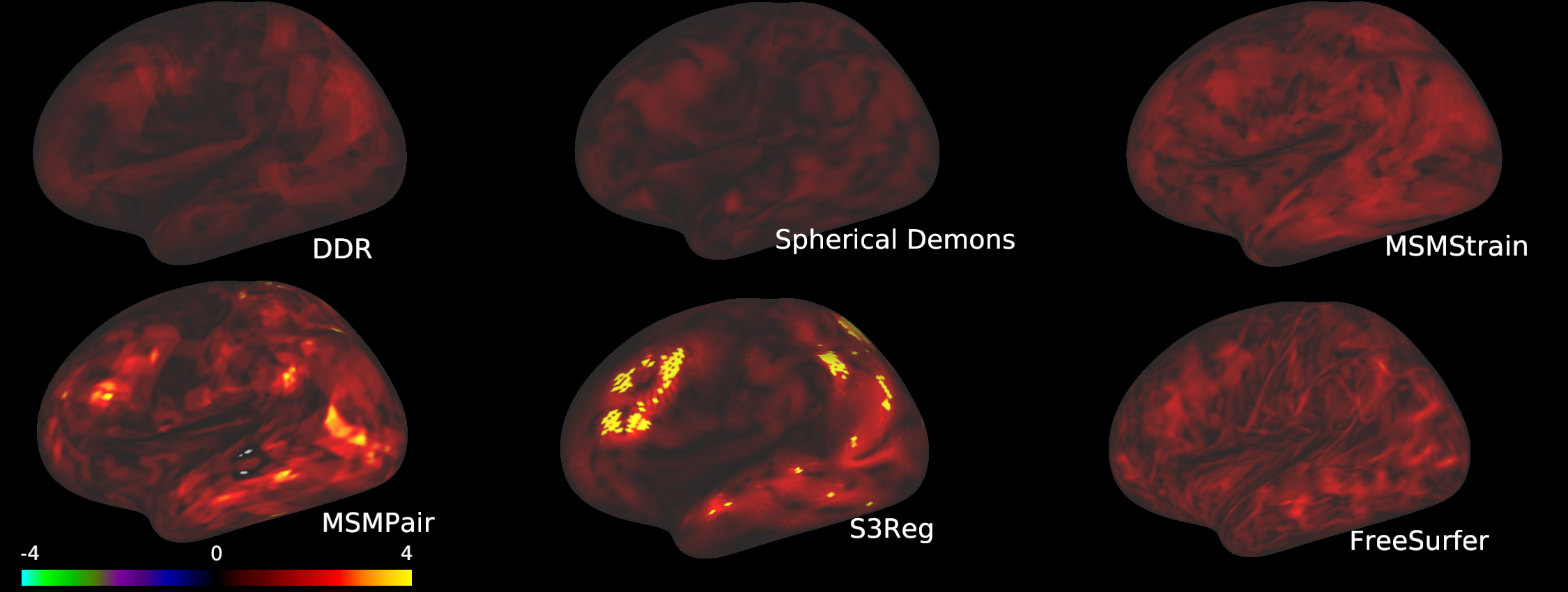}
         
         \caption{}
         \label{fig: subject b}
     \end{subfigure}
     \caption{(a), (c), (d) Qualitative comparisons of methods' performance on subjects with atypical cortical folding patterns at CC of $\sim 0.88$. (b), (d), (f) Associated Strain $J$ (top) and Strain $R$ (bottom).}
     \label{fig: reg subjects more}
\end{figure}

In Fig.~\ref{fig: reg subjects more}, we provide qualitative comparisons of the methods registration performance with the template on subjects with atypical cortical folding patterns at CC of $\sim 0.88$ (except for FreeSurfer, which has $0.75$ CC). Moreover, we also provide the associated strain $J$  and strain $R$ of these methods. The figure shows clearly that DDR, SD, and MSM Strain all produce good comparable alignments with reduced distortions, while MSM Pair and S3Reg provide alignments with peaks of high distortions.

\begin{figure}[h!]
     \centering
     \begin{subfigure}[b]{0.45\textwidth}
         \centering
         \includegraphics[width=\textwidth]{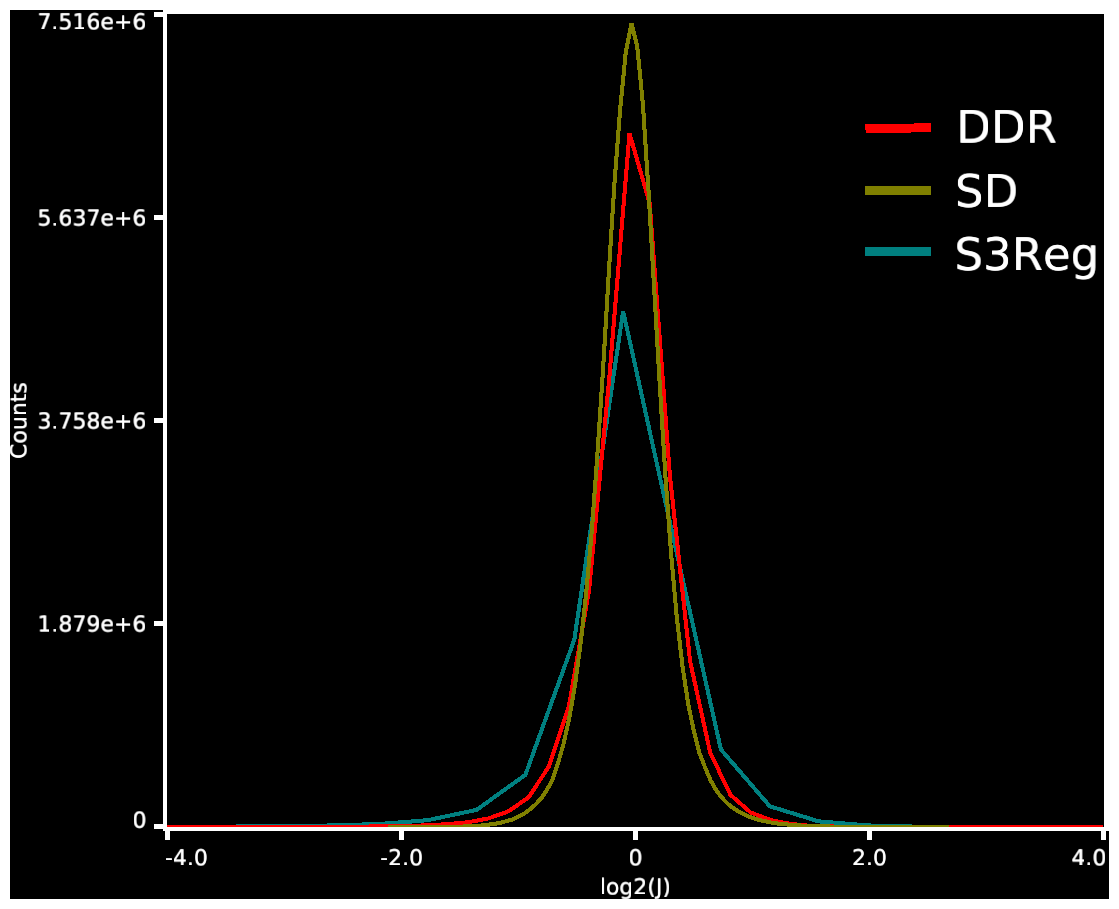}
         \caption{Strain $J$.}
         \label{fig: strainj0.9}
     \end{subfigure}
     \begin{subfigure}[b]{0.45\textwidth}
         \centering
         \includegraphics[width=\textwidth]{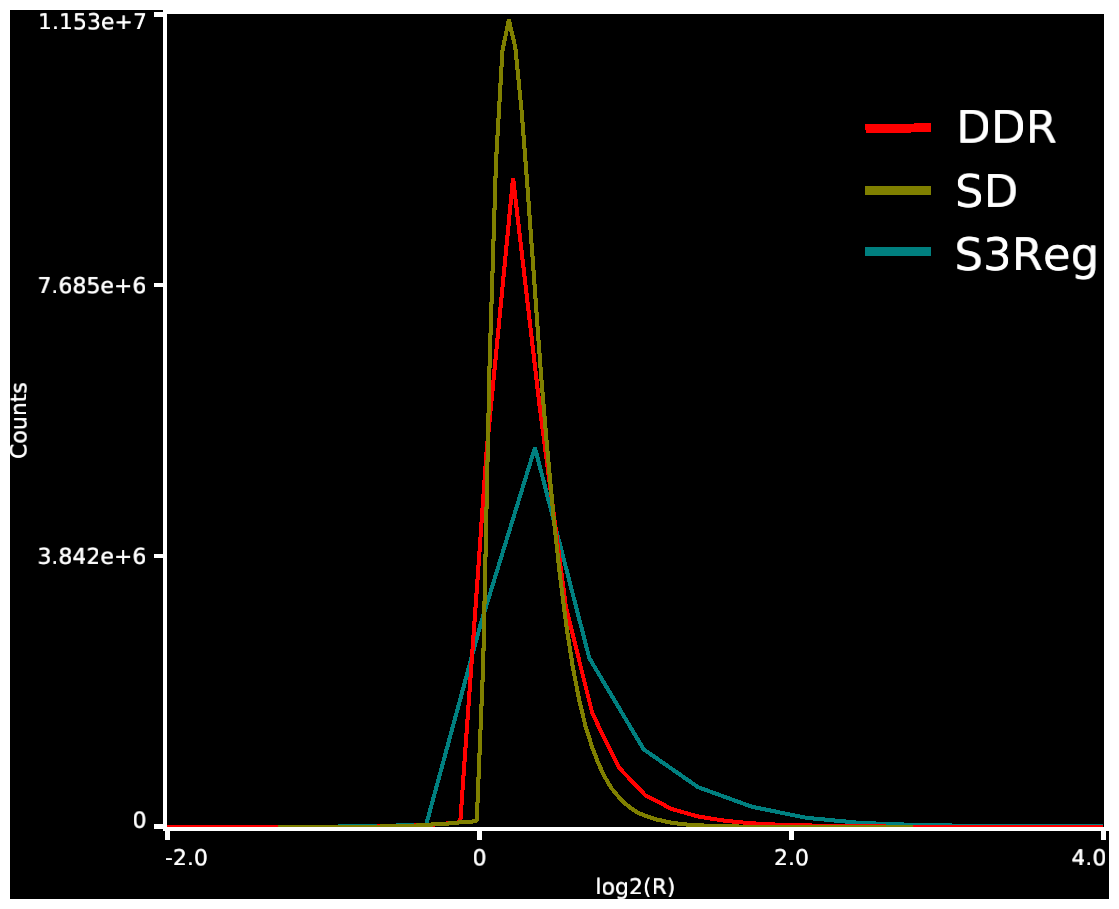}
         \caption{Strain $R$.}
         \label{fig: strainj0.9}
     \end{subfigure}
     \caption{Histogram plots comparing distortions across all test subjects for methods with CC $\sim  0.90$.}
     \label{fig: hist plot0.9}
\end{figure}

Finally, we plot in Fig.~\ref{fig: hist plot0.9} the histograms for $J$ and $R$ across all test subjects for methods that return a CC value of approximately $0.90$ (i.e., SD $= 0.901$ CC, DDR $= 0.895$ CC, and S3Reg $= 0.897$ CC). Fig.~\ref{fig: hist plot0.9} shows that most SD and DDR distortions at this CC level are around zero, while S3Reg leads to excess distortions.

\section{Conclusions}
In this work, we propose the first deep-discrete registration (DDR) framework for cortical surface alignment that aligns two surfaces by deforming a set of control points on the surface into a finite set of possible deformations. Our results show that DDR outperforms other deep learning-based cortical surface registration frameworks (S3Reg) in terms of similarity and distortions measures and provides a competitive performance to state-of-the-art conventional surface registration methods. Future work will extend DDR to multimodal alignment with high dimensional feature sets and address topographical variation.

\newpage


\begin{thebibliography}{10}
\providecommand{\url}[1]{\texttt{#1}}
\providecommand{\urlprefix}{URL }
\providecommand{\doi}[1]{https://doi.org/#1}

\bibitem{amunts2000brodmann}
Amunts, K., Malikovic, A., Mohlberg, H., Schormann, T., Zilles, K.: Brodmann's
  areas 17 and 18 brought into stereotaxic space—where and how variable?
  Neuroimage  \textbf{11}(1),  66--84 (2000)

\bibitem{aoki2019pointnetlk}
Aoki, Y., Goforth, H., Srivatsan, R.A., Lucey, S.: Pointnetlk: Robust \&
  efficient point cloud registration using pointnet. In: Proceedings of the
  IEEE/CVF Conference on Computer Vision and Pattern Recognition. pp.
  7163--7172 (2019)

\bibitem{balakrishnan2019voxelmorph}
Balakrishnan, G., Zhao, A., Sabuncu, M.R., Guttag, J., Dalca, A.V.: Voxelmorph:
  a learning framework for deformable medical image registration. IEEE
  transactions on medical imaging  \textbf{38}(8),  1788--1800 (2019)

\bibitem{borovec2020anhir}
Borovec, J., Kybic, J., Arganda-Carreras, I., Sorokin, D.V., Bueno, G.,
  Khvostikov, A.V., Bakas, S., Eric, I., Chang, C., Heldmann, S., et~al.:
  Anhir: automatic non-rigid histological image registration challenge. IEEE
  transactions on medical imaging  \textbf{39}(10),  3042--3052 (2020)

\bibitem{coalson2018impact}
Coalson, T.S., Van~Essen, D.C., Glasser, M.F.: The impact of traditional
  neuroimaging methods on the spatial localization of cortical areas.
  Proceedings of the National Academy of Sciences  \textbf{115}(27),
  E6356--E6365 (2018)

\bibitem{dalca2019learning}
Dalca, A., Rakic, M., Guttag, J., Sabuncu, M.: Learning conditional deformable
  templates with convolutional networks. Advances in neural information
  processing systems  \textbf{32} (2019)

\bibitem{de2019deep}
De~Vos, B.D., Berendsen, F.F., Viergever, M.A., Sokooti, H., Staring, M.,
  I{\v{s}}gum, I.: A deep learning framework for unsupervised affine and
  deformable image registration. Medical image analysis  \textbf{52},  128--143
  (2019)

\bibitem{fan2018adversarial}
Fan, J., Cao, X., Xue, Z., Yap, P.T., Shen, D.: Adversarial similarity network
  for evaluating image alignment in deep learning based registration. In:
  International Conference on Medical Image Computing and Computer-Assisted
  Intervention. pp. 739--746. Springer (2018)

\bibitem{fawaz2021benchmarking}
Fawaz, A., Williams, L.Z., Alansary, A., Bass, C., Gopinath, K., da~Silva, M.,
  Dahan, S., Adamson, C., Alexander, B., Thompson, D., et~al.: Benchmarking
  geometric deep learning for cortical segmentation and neurodevelopmental
  phenotype prediction. bioRxiv  (2021)

\bibitem{Fey/Lenssen/2019}
Fey, M., Lenssen, J.E.: Fast graph representation learning with {PyTorch
  Geometric}. In: ICLR Workshop on Representation Learning on Graphs and
  Manifolds (2019)

\bibitem{fischl1999high}
Fischl, B., Sereno, M.I., Tootell, R.B., Dale, A.M.: High-resolution
  intersubject averaging and a coordinate system for the cortical surface.
  Human brain mapping  \textbf{8}(4),  272--284 (1999)

\bibitem{fu2020lungregnet}
Fu, Y., Lei, Y., Wang, T., Higgins, K., Bradley, J.D., Curran, W.J., Liu, T.,
  Yang, X.: Lungregnet: An unsupervised deformable image registration method
  for 4d-ct lung. Medical physics  \textbf{47}(4),  1763--1774 (2020)

\bibitem{glasser2016multi}
Glasser, M.F., Coalson, T.S., Robinson, E.C., Hacker, C.D., Harwell, J.,
  Yacoub, E., Ugurbil, K., Andersson, J., Beckmann, C.F., Jenkinson, M.,
  et~al.: A multi-modal parcellation of human cerebral cortex. Nature
  \textbf{536}(7615),  171--178 (2016)

\bibitem{glasser2013minimal}
Glasser, M.F., Sotiropoulos, S.N., Wilson, J.A., Coalson, T.S., Fischl, B.,
  Andersson, J.L., Xu, J., Jbabdi, S., Webster, M., Polimeni, J.R., et~al.: The
  minimal preprocessing pipelines for the human connectome project. Neuroimage
  \textbf{80},  105--124 (2013)

\bibitem{heinrich2019closing}
Heinrich, M.P.: Closing the gap between deep and conventional image
  registration using probabilistic dense displacement networks. In:
  International Conference on Medical Image Computing and Computer-Assisted
  Intervention. pp. 50--58. Springer (2019)

\bibitem{heinrich2020highly}
Heinrich, M.P., Hansen, L.: Highly accurate and memory efficient unsupervised
  learning-based discrete ct registration using 2.5 d displacement search. In:
  International Conference on Medical Image Computing and Computer-Assisted
  Intervention. pp. 190--200. Springer (2020)

\bibitem{kingma2014adam}
Kingma, D.P., Ba, J.: Adam: A method for stochastic optimization. arXiv
  preprint arXiv:1412.6980  (2014)

\bibitem{krahenbuhl2011efficient}
Kr{\"a}henb{\"u}hl, P., Koltun, V.: Efficient inference in fully connected crfs
  with gaussian edge potentials. Advances in neural information processing
  systems  \textbf{24} (2011)

\bibitem{monti2017geometric}
Monti, F., Boscaini, D., Masci, J., Rodola, E., Svoboda, J., Bronstein, M.M.:
  Geometric deep learning on graphs and manifolds using mixture model cnns. In:
  Proceedings of the IEEE conference on computer vision and pattern
  recognition. pp. 5115--5124 (2017)

\bibitem{pielawski2020comir}
Pielawski, N., Wetzer, E., {\"O}fverstedt, J., Lu, J., W{\"a}hlby, C.,
  Lindblad, J., Sladoje, N.: Comir: Contrastive multimodal image representation
  for registration. Advances in neural information processing systems
  \textbf{33},  18433--18444 (2020)

\bibitem{qi2017pointnet}
Qi, C.R., Su, H., Mo, K., Guibas, L.J.: Pointnet: Deep learning on point sets
  for 3d classification and segmentation. In: Proceedings of the IEEE
  conference on computer vision and pattern recognition. pp. 652--660 (2017)

\bibitem{robinson2018multimodal}
Robinson, E.C., Garcia, K., Glasser, M.F., Chen, Z., Coalson, T.S.,
  Makropoulos, A., Bozek, J., Wright, R., Schuh, A., Webster, M., et~al.:
  Multimodal surface matching with higher-order smoothness constraints.
  Neuroimage  \textbf{167},  453--465 (2018)

\bibitem{robinson2014msm}
Robinson, E.C., Jbabdi, S., Glasser, M.F., Andersson, J., Burgess, G.C., Harms,
  M.P., Smith, S.M., Van~Essen, D.C., Jenkinson, M.: Msm: a new flexible
  framework for multimodal surface matching. Neuroimage  \textbf{100},
  414--426 (2014)

\bibitem{shao2021prosregnet}
Shao, W., Banh, L., Kunder, C.A., Fan, R.E., Soerensen, S.J., Wang, J.B.,
  Teslovich, N.C., Madhuripan, N., Jawahar, A., Ghanouni, P., et~al.:
  Prosregnet: A deep learning framework for registration of mri and
  histopathology images of the prostate. Medical image analysis  \textbf{68},
  101919 (2021)

\bibitem{wang2019deep}
Wang, Y., Solomon, J.M.: Deep closest point: Learning representations for point
  cloud registration. In: Proceedings of the IEEE/CVF International Conference
  on Computer Vision. pp. 3523--3532 (2019)

\bibitem{yeo2009spherical}
Yeo, B.T., Sabuncu, M.R., Vercauteren, T., Ayache, N., Fischl, B., Golland, P.:
  Spherical demons: fast diffeomorphic landmark-free surface registration. IEEE
  transactions on medical imaging  \textbf{29}(3),  650--668 (2009)

\bibitem{zhao2021s3reg}
Zhao, F., Wu, Z., Wang, F., Lin, W., Xia, S., Shen, D., Wang, L., Li, G.:
  S3reg: superfast spherical surface registration based on deep learning. IEEE
  Transactions on Medical Imaging  \textbf{40}(8),  1964--1976 (2021)

\bibitem{zhao2019spherical}
Zhao, F., Xia, S., Wu, Z., Duan, D., Wang, L., Lin, W., Gilmore, J.H., Shen,
  D., Li, G.: Spherical u-net on cortical surfaces: methods and applications.
  In: International Conference on Information Processing in Medical Imaging.
  pp. 855--866. Springer (2019)

\bibitem{zheng2015conditional}
Zheng, S., Jayasumana, S., Romera-Paredes, B., Vineet, V., Su, Z., Du, D.,
  Huang, C., Torr, P.H.: Conditional random fields as recurrent neural
  networks. In: Proceedings of the IEEE international conference on computer
  vision. pp. 1529--1537 (2015)

\bibitem{zhou2019continuity}
Zhou, Y., Barnes, C., Lu, J., Yang, J., Li, H.: On the continuity of rotation
  representations in neural networks. In: Proceedings of the IEEE/CVF
  Conference on Computer Vision and Pattern Recognition. pp. 5745--5753 (2019)

\end{thebibliography}

%
%
%
%
%
%
\end{document}